\documentclass[10pt,twocolumn,letterpaper]{article}

\usepackage{cvpr}              

\usepackage{graphicx}
\usepackage{amsmath}
\usepackage{amssymb}
\usepackage{booktabs}

\usepackage[pagebackref,breaklinks,colorlinks]{hyperref}

\usepackage[capitalize]{cleveref}
\crefname{section}{Sec.}{Secs.}
\Crefname{section}{Section}{Sections}
\Crefname{table}{Table}{Tables}
\crefname{table}{Tab.}{Tabs.}

\def\cvprPaperID{4504} 
\def\confName{CVPR}
\def\confYear{2022}

\begin{document}

\title{Progressively Generating Better Initial Guesses Towards Next Stages for High-Quality Human Motion Prediction}
\author{Tiezheng Ma${}^1$, Yongwei Nie${}^{1{\thanks{Corresponding author: nieyongwei@scut.edu.cn}}}$, Chengjiang Long${}^2$, Qing Zhang${}^3$ and Guiqing Li${}^1$\\
\\
$^1{}${School of Computer Science and Engineering, South China University of Technology, China}\\
$^2{}${Meta Reality Labs, USA}\\
$^3{}$School of Computer Science and Engineering, Sun Yat-sen University, China
}

\maketitle

\begin{abstract}
   This paper presents a high-quality human motion prediction method that accurately predicts future human poses given observed ones. Our method is based on the observation that a good {\rm``initial guess''} of the future poses is very helpful in improving the forecasting accuracy. This motivates us to propose a novel two-stage prediction framework, including an init-prediction network that just computes the good guess and then a formal-prediction network that predicts the target future poses based on the guess. More importantly, we extend this idea further and design a multi-stage prediction framework where each stage predicts initial guess for the next stage, which brings more performance gain. To fulfill the prediction task at each stage, we propose a network comprising Spatial Dense Graph Convolutional Networks (S-DGCN) and Temporal Dense Graph Convolutional Networks (T-DGCN). Alternatively executing the two networks helps extract spatiotemporal features over the global receptive field of the whole pose sequence. All the above design choices cooperating together make our method outperform previous approaches by large margins: 6\%-7\% on Human3.6M, 5\%-10\% on CMU-MoCap, and 13\%-16\% on 3DPW. Code is available at \href{https://github.com/705062791/PGBIG}{https://github.com/705062791/PGBIG}.
\end{abstract}

\section{Introduction}
\label{sec:intro}

\begin{figure} [!t]
    \centering
    \includegraphics[width=\columnwidth]{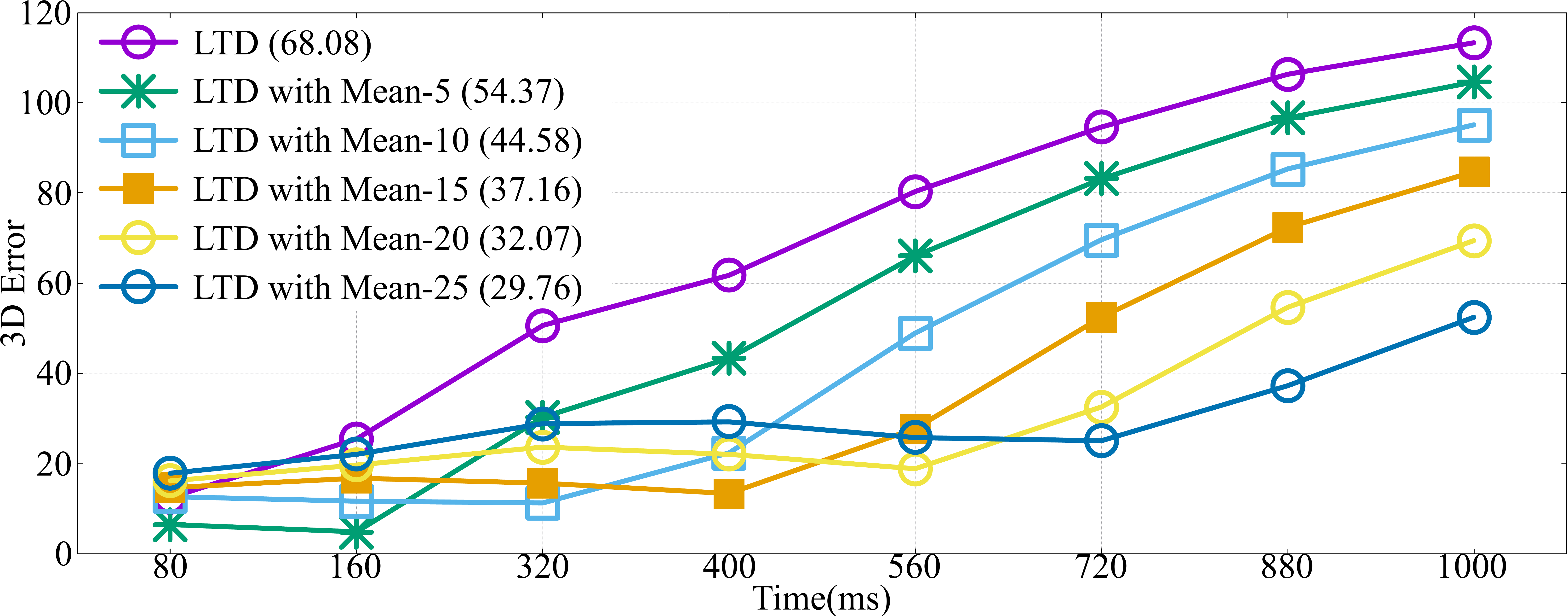} \\
    (a)\\
    \includegraphics[width=0.95\columnwidth]{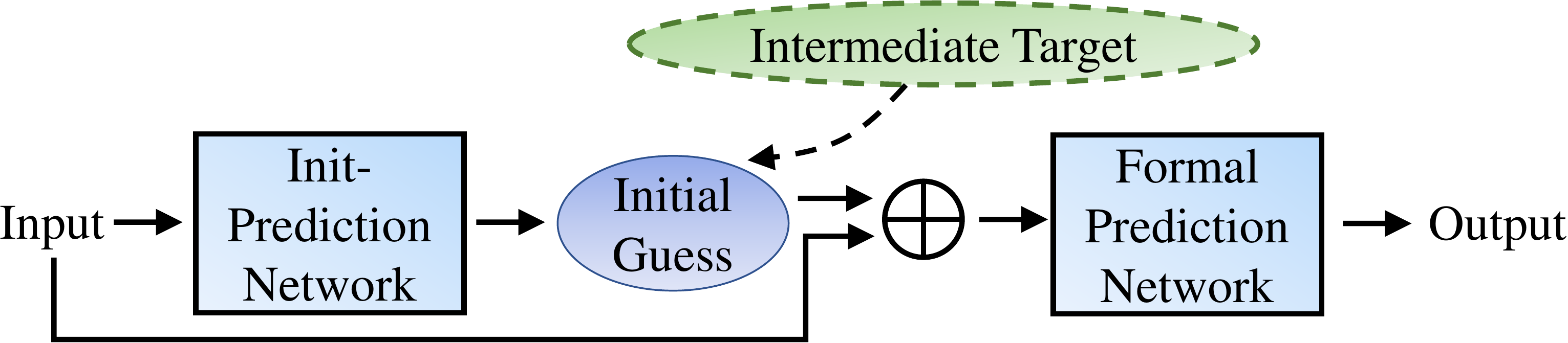}\\
    (b)
    \vspace{-0.3cm}
    \caption{(a) Toy experiments. Given 10 poses, we predict 25 poses. The frame rate is 25fps, and 25 poses last 1000ms. We use LTD~\cite{mao2019learning} as the baseline which uses the last observed pose as the initial guess. At test time, the average prediction error of LTD is 68.08. We conduct another 5 experiments training and testing LTD with Mean-$x$ as the initial guess. That is, Mean-$x$ is duplicated and appended to the past poses where ``Mean-$x$'' is the mean of the first future $x$ poses with $x$ belonging to $\{5,10,15,20,25\}$. As $x$ increases, the average prediction error significantly decreases, meaning that when used as initial guess, ``Mean-$x$'' is better than the last observed pose, and the larger the $x$ the better. The curves in the figure plot the prediction error at every forecasting time.
    (b) Our two-stage prediction framework comprising an init-prediction network and a formal-prediction network. The init-prediction network is supervised by an intermediate target.}
    \label{fig:teaser}
\end{figure}

Human Motion Prediction (HMP) is a fundamental research topic that benefits many other applications such as intelligent security, autonomous driving, human-robot interaction and so on. Early works employed nonlinear Markov models \cite{lehrmann2014efficient}, Gaussian Process dynamical models \cite{wang2005gaussian}, and Restricted Boltzmann Machine \cite{taylor2007modeling} to tackle this problem, while recently a large number of methods based on deep learning have emerged, showing significant merits. 
 
Due to the sequential nature of pose sequences, HMP is mostly tackled with Recurrent Neural Networks (RNN) \cite{fragkiadaki2015recurrent,jain2016structural,ghosh2017learning,martinez2017human,gui2018adversarial,tang2018long,gui2018few,guo2019human,liu2019towards,chiu2019action,gopalakrishnan2019neural,sang2020human,corona2020context,pavllo2020modeling}. However, RNN-based approaches usually yield problems of discontinuity and error accumulation which might be due to the training difficulty of RNNs. There are a few works that employ Convolutional Neural Networks (CNN) to solve the HMP problem \cite{butepage2017deep,li2018convolutional,shu2021spatiotemporal,cui2021efficient}. They treat a pose sequence as an image and apply 2D convolutions to the pose sequence, but poses are essentially not regular data which limits the effectiveness of the 2D convolutions. Recently, lots of works demonstrate that Graph Convolutional Networks (GCN) is very suitable for HMP \cite{aksan2019structured,mao2019learning,mao2020history,cui2020learning,li2020dynamic,li2021symbiotic,li2020multitask,liu2020multi,lebailly2020motion,dang2021msr,cui2021towards}. They treat a human pose as a graph by viewing each joint as a node of the graph and constructing edges between any pair of joints. GCNs are then used to learn spatial relations between joints which benefit the pose prediction. 

We observe that starting from the seminal work of LTD~\cite{mao2019learning}, all recent GCN-based approaches \cite{dang2021msr,cui2020learning,mao2020history,sofianos2021space} share the following preprocessing steps: (1) They duplicate the last observed pose as many times as the length of the future pose sequence, and append the duplicated poses to the observed sequence to form an extended input sequence. (2) Similarly, the ground truth future poses are appended to the observed poses to obtain the extended ground truth output sequence. Their proposed networks are used to predict from the extended input sequence to the extended output sequence instead of from the original observed poses to the future poses. Ablation comparisons show that the prediction between the extended sequences is easier than between the original sequences, and the former achieves significantly better prediction accuracy than the latter. Dang \etal \cite{dang2021msr} ascribed this to the global residual connection between the extended input and output, while in this paper we interpret this phenomenon from another perspective: the last observed pose provides an \textit{``initial guess''} for the target future poses. From the initial guess, the network just needs to move slightly such that it can reach the target positions. However, we argue that the last observed pose is not the best initial guess. For example, the toy experiments in Figure \ref{fig:teaser} (a) show that the mean pose of future poses is better than the last observed pose as the initial guess.

The problem is that we do not really know the mean pose of the future poses. Thus as shown in Figure~\ref{fig:teaser} (b), using the mean of future poses as intermediate target, we propose to predict the mean of the future poses firstly and then predict the final target future poses by viewing the \textit{predicted mean} as the initial guess. Although the predicted mean is not as good as the ground truth mean when used as the initial guess, it is better than the last observed pose. Further, for more accuracy gain, we extend the two-stage prediction strategy to a multi-stage version. To this end, we recursively smooth the ground truth output sequence, obtaining a set of sequences at different smoothing levels. By treating these smoothed results as intermediate targets at the multiple stages, our multi-stage prediction framework progressively predicts better initial guesses towards the next stages until the final target pose sequence obtained.

Any existing human motion prediction model such as \cite{martinez2017human,li2018convolutional,mao2019learning} can be used to accomplish the prediction task at each of our stages. Among them, we choose GCN as the buildingblock to construct our multi-stage framework. Existing GCN-based approaches \cite{mao2019learning,cui2020learning,dang2021msr} only employ GCN to extract spatial features.
Instead of them, we propose to process both spatial and temporal features by GCNs. Specifically, we propose S-DGCN and T-DGCN. S-DGCN views each pose as a fully-connected graph and encodes global spatial dependencies in human pose, while T-DGCN views each joint trajectory as a fully-connected graph and encodes global temporal dependencies in motion trajectory. S-DGCN and T-DGCN together extract global spatiotemporal features, which further improve our prediction accuracy.

In summary, the main contributions of this paper are three-fold: 
\begin{itemize}
    \item We propose a novel multi-stage human motion prediction framework utilizing recursively smoothed results of the ground truth target sequence as the intermediate targets, by which we progressively improve the initial guess of the final target future poses for better prediction accuracy.
    \item We propose a network based on S-DGCN and T-DGCN that extracts global spatiotemporal features effectively to fulfill the prediction task at each stage.
    \item We conduct extensive experiments showing that our method outperforms previous approaches by large margins on three public datasets.
\end{itemize}

\begin{figure*}[ht]
    \centering
    \includegraphics[width=1\textwidth]{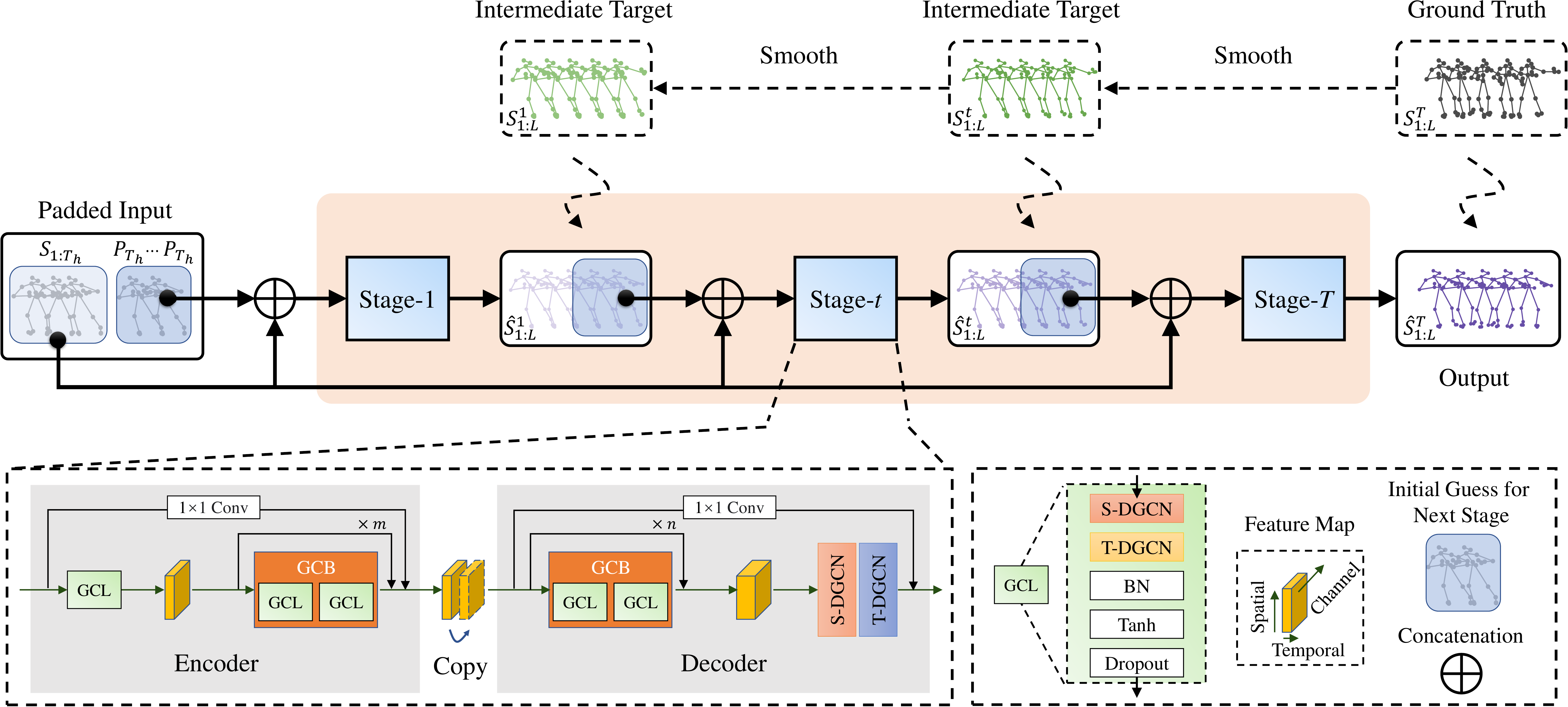}\\
    \vspace{-0.3cm}
    \caption{Overview of our multi-stage human motion prediction framework containing $T$ stages. Each stage takes the observed sequence $S_{1:T_h}$ and an initial guess as input. For the first stage, the initial guess is composed of the last observed pose. For all the other stages, the initial guess is the future part of the output of previous stage. The last stage is guided by the ground truth, while all the other stages are guided by the corresponding recursively smoothed results of the ground truth. All the stages use the same Encoder-Copy-Decoder prediction network. Please refer to the main text for more details.}
    \label{fig:overview}
\end{figure*}

\section{Related Work}

Due to the serialized nature of human motion data, most previous works adopt RNN as backbone \cite{fragkiadaki2015recurrent,jain2016structural,ghosh2017learning,martinez2017human,gui2018adversarial,tang2018long,gui2018few,guo2019human,liu2019towards,chiu2019action,gopalakrishnan2019neural,sang2020human,corona2020context,pavllo2020modeling}. For example, ERD \cite{fragkiadaki2015recurrent} improves the recurrent layer of LSTM \cite{hochreiter1997long} by placing an encoder before it and a decoder after it. Jain \etal \cite{jain2016structural} organized RNNs according to the spatiotemporal structure of human pose, proposing the Structural-RNN. Martinez \etal \cite{martinez2017human} used sequence to sequence architecture that is often adopted for language processing to predict human motion. RNNs are hard to train and cannot effectively capture spatial relationships between joints, usually yielding problems of discontinuity and error accumulation.

To enhance the ability of extracting spatial features of human pose, Shu \etal \cite{shu2021spatiotemporal} compensated RNN with skeleton-joint co-attention mechanism. The works of \cite{butepage2017deep,li2018convolutional,liu2020trajectorycnn} use CNNs for this purpose but CNNs cannot directly model the interaction between any pair of joints.

Viewing human pose as a graph, recent works have popularly adopted GCNs for human motion prediction \cite{aksan2019structured,mao2019learning,mao2020history,cui2020learning,li2020dynamic,li2021symbiotic,li2020multitask,liu2020multi,lebailly2020motion,dang2021msr,cui2021towards,Shi:AAAI2022,Shi:CVPR2021,Duan:AAAI2022}. Aksan \etal \cite{aksan2019structured} did not use GCN, but they adopted a very similar idea that relies on many small networks to exchange features between adjacent joints. The works of \cite{li2020dynamic,li2021symbiotic,lebailly2020motion} use GCN either in the encoder \cite{li2020dynamic,li2021symbiotic} for feature encoding or in the decoder \cite{lebailly2020motion} for better decoding. The works of \cite{mao2019learning,mao2020history,cui2020learning,dang2021msr} are totally based on GCN. Mao \etal \cite{mao2019learning} viewed a pose as a fully-connected graph and used GCN to discover the relationship between any pair of joints. In the temporal domain, they represented the joint trajectories by Discrete Cosine Transform coefficients. Dang \etal \cite{dang2021msr} extended \cite{mao2019learning} to a multi-scale version across the abstraction levels of human pose. We also use GCN as the basic buildingblock, but propose S-DGCN and T-DGCN that extract global spatiotemporal features, better than \cite{mao2019learning,mao2020history,dang2021msr} that just extract spatial features. Recently, Sofianos \etal \cite{sofianos2021space} proposed a method that can also extract spatiotemporal features by GCNs. The difference is that we achieve that by only two GCNs while \cite{sofianos2021space} uses much more GCNs.

Transformer \cite{vaswani2017attention,Dong:MM2021} has also been adapted to tackle the problem of human motion prediction \cite{aksan2020spatio,cai2020learning}. Similar to GCN, the self-attention mechanism of Transformer can compute pairwise relations of joints. In this paper, we choose GCN as the buildingblock. We show that our proposed method outperforms the existing Transformer-based approaches in terms of both running time and accuracy.

\section{Methodology}
Let $S_{1:T_h} = \{P_1,P_2,\cdots,P_{T_h}\}$ denote an observed pose sequence of length $T_h$ where $P_i$ is a pose at time $i$, and $S_{T_h+1:T_h+T_f}$ be the future pose sequence of length $T_f$. Instead of directly mapping from $S_{1:T_h}$ to $S_{T_h+1:T_h+T_f}$, we follow \cite{mao2019learning,mao2020history,dang2021msr} to repeat the last observed pose $P_{T_h}$, $T_f$ times and append them to $S_{1:T_h}$, obtaining the padded input sequence $[S_{1:T_h};P_{T_h},\cdots,P_{T_h}]$ of length $L$ with $L=T_h+T_f$. Then our aim becomes to find a mapping from the padded sequence to its ground truth $S_{1:L}=[S_{1:T_h};S_{T_h+1:T_h+T_f}]$.

\subsection{Multi-Stage Progressive Prediction Framework}

For the above purpose, we design a multi-stage progressive prediction framework as shown in Figure~\ref{fig:overview} (the two-stage framework shown in Figure~\ref{fig:teaser} (b) is a special case of the multi-stage framework), which contains $T$ stages represented by $\Phi^1,\Phi^2,\cdots,\Phi^T$ respectively. These stages perform the following subtasks step by step:
\begin{equation}
\begin{aligned}
     \hat{S}_{1:L}^{1} &= \Phi^1([S_{1:T_h};P_{T_h},\cdots,P_{T_h}]), \\
    \hat{S}_{1:L}^{i} &= \Phi^i([S_{1:T_h};\hat{S}_{T_h+1:L}^{i-1}]), i = {2,3,\cdots,T},
\end{aligned}
\end{equation}
in which $\hat{S}_{1:L}^{i}$ is the output of stage $i$. The input to every stage is composed of two parts: the observed poses $S_{1:T_h}$ and the initial guess. For the first stage, the initial guess is $P_{T_h},\cdots,P_{T_h}$. For stage $i$, the initial guess is $\hat{S}_{T_h+1:L}^{i-1}$ which is the future part of the output at the previous stage.

Recall that for the two-stage prediction framework as shown in Figure~\ref{fig:teaser} (b), the mean pose of future poses is used as the intermediate target, while for the multi-stage framework we resort to smoothing $S^{T}_{1:L}$ ($=S_{1:L}$) recursively to obtain $S^{T-1}_{1:L},S^{T-2}_{1:L},\cdots,S^{1}_{1:L}$, and use them as the intermediate targets of the corresponding stage networks $\Phi^T,\Phi^{T-1},\cdots,\Phi^1$ to guide the generation of $\hat{S}_{1:L}^{T},\hat{S}_{1:L}^{T-1},\cdots,\hat{S}_{1:L}^{1}$ (in reverse order), respectively. The adopted smoothing algorithm is Accumulated Average Smoothing (AAS) which is introduced in the following.

Let each pose have $M$ joints, and each joint be a point in the $D$-dimensional space. For a pose sequence $S^{T}_{1:L}$, we have $M\times D$ trajectories: $\{T_j|j\in[1, M\times D]\}$, and each trajectory $T_j$ is composed of the same coordinate across all the poses: $T_j=\{x^i_j|i\in[1,L]\}$. Since all of the trajectories are smoothed by the same method, we omit the subscript $j$ in the following without loss of generality.

Note that the trajectory contains two parts: the historical part $\{x^i|i\in[1,T_h]\}$ and the future part $\{x^i|i\in[T_h+1,T_h+T_f]\}$. We just need to smooth the future part and keep the historical part unchanged. The AAS algorithm is defined as:
\begin{equation}
    \bar{x}^i = \frac{1}{i-T_h}\sum_{k=T_h+1}^{i}x^k, \forall i\in[T_h+1, T_h+T_f].
\end{equation}
That is, the smoothed value of a point on a curve is the average of all the previous points on the curve. We apply AAS to $S^{T}_{1:L}$ recursively, obtaining $S^{T-1}_{1:L},S^{T-2}_{1:L},\cdots,S^{1}_{1:L}$.

\begin{figure}[!t]
    \centering
    \includegraphics[width=1.0\columnwidth]{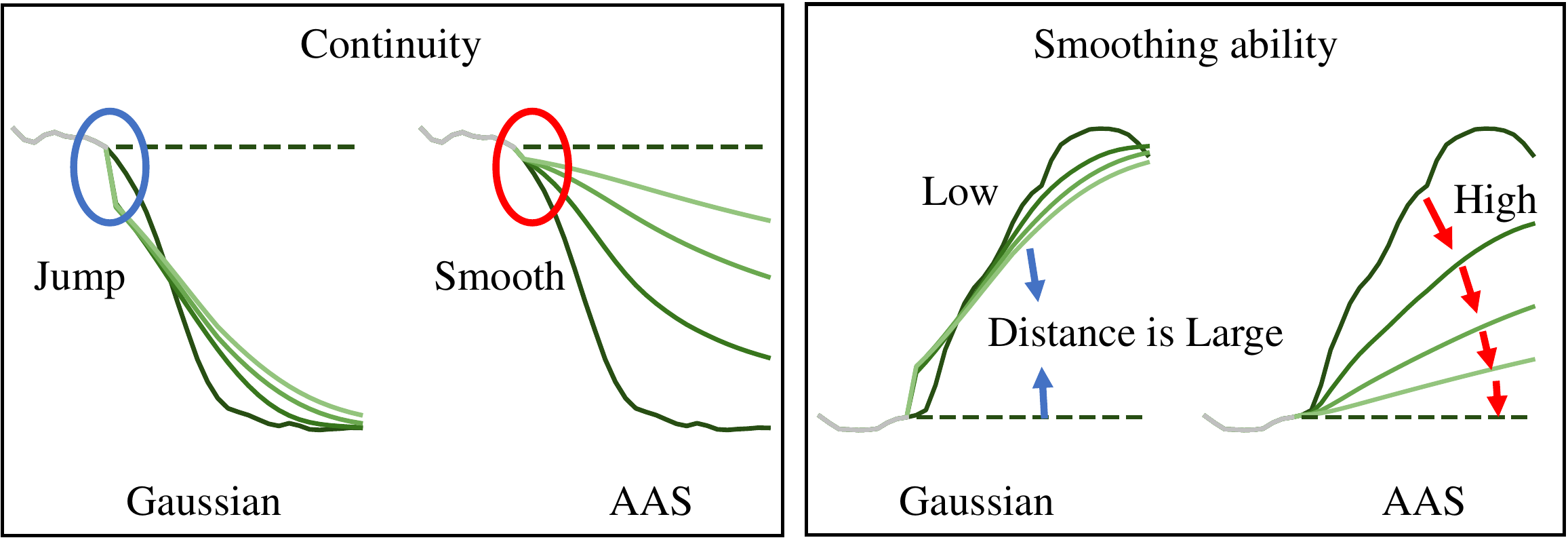}\\
    \vspace{-0.3cm}
    \caption{Comparisons between AAS and Gaussian filter on recursively smoothing effects. In each figure, left shows results of Gaussian filter, right shows results of AAS. The gray curve indicates a historical trajectory, the black is the ground truth future trajectory, and the curves from dark to light blue are recursively smoothed results. The left figure shows that AAS keeps the continuity between the historical and smoothed curves while Gaussian filter does not. The right figure shows that AAS has stronger smoothing ability than Gaussian filter.}
    \label{fig:aas-and-gau}
\end{figure}

Figure \ref{fig:aas-and-gau} shows results by AAS and compares them with those by a Gaussian filter (standard normal distribution) with filtering window size of 21. In each group of curves, the gray curve represents a historical trajectory, the black is the ground truth trajectory in the future, and the dash line is obtained by padding the last observed data. From dark to light blue are the recursively smoothed results. Compared with Gaussian filter, AAS has two advantages. (1) AAS preserves the continuity between the historical and future trajectories, while Gaussian filter yields jumps at the junctions. (2) AAS has stronger smoothing ability than Gaussian filter. As can be seen, the results by AAS evenly and steadily approach the dash line. The dash line is a good guess of the smoothest curve of AAS. Meanwhile, each curve by AAS is a good guess of the curve at the previous smoothing level. From this point, AAS is very suitable for preparing intermediate targets for our multi-stage framework. In contrast, the results of Gaussian filter are concentrated together, and all of them are far from the dash line.


\subsection{Encoder-Copy-Decoder Stage Prediction Network Comprising S-DGCN and T-DGCN}

In this section, we introduce our network that fulfills the prediction task at each stage, the overview of which is illustrated at the bottom-left of Figure~\ref{fig:overview}. Our network is totally based on GCNs. Specifically, we propose S-DGCN and T-DGCN that extract global spatial and temporal interactions between joints. Based on S-DGCN and T-DGCN, we build an Encoder-Copy-Decoder prediction network. In the following, we introduce them one by one.




\textbf{S-DGCN.} By Dense GCN, \ie DGCN, we mean the processed graph is fully connected. S-DGCN defines a spatially dense graph convolution applied to a pose, and the graph convolution is shared by all the poses of a pose sequence. Let $X\in \mathbb{R}^{L\times M \times F}$ be a pose sequence where $L$ is the length of the sequence, $M$ is the number of joints of a pose, and $F$ indicates the number of features of a joint. Defining a learnable adjacency matrix $A^s\in\mathbb{R}^{M\times M}$ the elements of which measure relationships between pairs of joints of a pose, S-DGCN computes:
\begin{equation}
X^\prime = \text{S-DGCN}(X) = A^s X W^s,
\end{equation}
where $W^s\in\mathbb{R}^{F\times F^\prime}$ indicates the learnable parameters of S-DGCN, and $X^\prime\in\mathbb{R}^{L\times M\times F^\prime}$ is the output of S-DGCN.

\textbf{T-DGCN.} T-DGCN defines a temporal graph convolution applied to a joint trajectory, and the graph convolution is shared by all the trajectories. We first transpose the first two dimensions of $X^\prime$ to obtain $Y\in\mathbb{R}^{M\times L \times F^\prime}$. Defining a learnable adjacency matrix $A^t \in \mathbb{R}^{L\times L}$ measuring weights between pairs of joints of a trajectory, T-DGCN computes:
\begin{equation}
Y^\prime = \text{T-DGCN}(Y) = A^t Y W^t,
\end{equation}
where $W^t \in \mathbb{R}^{F^\prime\times F^\prime}$ is the learnable parameters of T-DGCN, and $Y^\prime \in \mathbb{R}^{M\times L\times F^\prime}$. Finally, we transpose the first two dimensions back to make $Y^\prime \in \mathbb{R}^{L\times M\times F^\prime}$.

\textbf{GCL.} As shown at the bottom-right of Figure~\ref{fig:overview}, we define a Graph Convolutional Layer (GCL) as a unit that sequentially executes S-DGCN, T-DGCN, batch normalization \cite{ioffe2015batch}, tanh, and dropout \cite{srivastava2014dropout}. GCL can extract spatiotemporal features over the global receptive field of the whole pose sequence.

\textbf{Encoder.} As shown in Figure \ref{fig:overview}, the encoder is a residual block containing a GCL and multiple Graph Convolutional Blocks (GCB). The first GCL projects the input from the pose space of $\mathbb{R}^{L\times M\times D}$ to the feature space of $\mathbb{R}^{L\times M\times F}$. We set $F=16$ in this paper. Each GCB is a residual block containing two GCLs. They always work in the feature space. In order to add the global residual connection for the encoder, we employ a $1\times 1$ convolutional layer with 16 kernels that maps the input into the space of $\mathbb{R}^{L\times M\times F}$ which is then added to the output of the GCBs.

\textbf{Copy.} The encoder outputs a feature map in the space of $\mathbb{R}^{L\times M\times F}$. We duplicate it and append the copy to the original feature map along the trajectory direction, obtaining a feature map of size $\mathbb{R}^{2L\times M\times F}$ which is used as the input to the decoder. We find in practice that the ``copy'' operator improves the prediction performance. The effectiveness of ``copy'' can be intuitively explained by the fact that the ``copy'' operator doubles the size of the latent space, enabling more parameters in the decoder to ensure more sufficient feature fusing.

\textbf{Decoder.} The decoder is a residual block containing multiple GCBs and a pair of S-DGCN and T-DGCN. The GCBs work in the feature space of $F=16$, while the final S-DGCN and T-DGCN project the features back into the pose space. Since the input to the decoder is of length $2L$, the adjacency matrix $A^t$ of all the T-DGCNs, including those in the GCBs, are of size $\mathbb{R}^{2L\times 2L}$. In order to add the residual connection for the decoder, a $1\times 1$ convolutional layer with 3 kernels is applied to the input of the decoder. The result of the decoder is of length $2L$, while we just retain the front $L$ poses as the final result.

\subsection{Loss Function} 

We apply $\mathcal{L}_1$ loss on all the outputs: $L =  \sum_{i=1}^{T}\|\hat{S}_{1:L}^{i}-S_{1:L}^{i} \|^2$.

\begin{table*}[ht]
\vspace{-0.3cm}
\caption{Comparisons of short-term prediction on Human3.6M. Results at 80ms, 160ms, 320ms, 400ms in the future are shown. The best results are highlighted in bold, and the second best are marked by underline.}
\vspace{-0.3cm}
\renewcommand\arraystretch{0.9}
\resizebox{\textwidth}{!}{
\begin{tabular}{c|cccc|cccc|cccc|cccc} \hline
\textbf{scenarios} & \multicolumn{4}{c|}{walking}                                   & \multicolumn{4}{c|}{eating}                                    & \multicolumn{4}{c|}{smoking}                                   & \multicolumn{4}{c}{discussion}                                \\ \hline
millisecond        & 80ms          & 160ms         & 320ms         & 400ms         & 80ms          & 160ms         & 320ms         & 400ms         & 80ms          & 160ms         & 320ms         & 400ms         & 80ms          & 160ms         & 320ms         & 400ms         \\ \hline
Res. Sup.          & 29.4          & 50.8          & 76.0          & 81.5          & 16.8          & 30.6          & 56.9          & 68.7          & 23.0          & 42.6          & 70.1          & 82.7          & 32.9          & 61.2          & 90.9          & 96.2          \\
DMGNN              & 17.3          & 30.7          & 54.6          & 65.2          & 11.0          & 21.4          & 36.2          & 43.9          & 9.0           & 17.6          & 32.1          & 40.3          & 17.3          & 34.8          & 61.0          & 69.8          \\
LTD                & 12.3          & 23.0          & 39.8          & 46.1          & 8.4           & \underline{16.9}          & 33.2          & 40.7          & \underline{7.9}           & \underline{16.2}          & 31.9          & 38.9          & 12.5          & 27.4          & 58.5          & 71.7          \\
MSR                & \underline{12.2}          & \underline{22.7}          & \underline{38.6}          & \underline{45.2}          & \underline{8.4}           & 17.1          & \underline{33.0}          & \underline{40.4}          & 8.0           & 16.3          & \underline{31.3}          & \underline{38.2}          & \underline{12.0}          & \underline{26.8}          & \underline{57.1}          & \underline{69.7}          \\
Ours                & \textbf{10.2}          & \textbf{19.8}          & \textbf{34.5}          & \textbf{40.3}          & \textbf{7.0}           & \textbf{15.1}          & \textbf{30.6}          & \textbf{38.1}          & \textbf{6.6}           & \textbf{14.1}          & \textbf{28.2}          & \textbf{34.7}          & \textbf{10.0}          & \textbf{23.8}          & \textbf{53.6}          & \textbf{66.7}          \\ \hline
scenarios          & \multicolumn{4}{c|}{directions}                                & \multicolumn{4}{c|}{greeting}                                  & \multicolumn{4}{c|}{phoning}                                   & \multicolumn{4}{c}{posing}                                    \\ \hline
millisecond        & 80ms          & 160ms         & 320ms         & 400ms         & 80ms          & 160ms         & 320ms         & 400ms         & 80ms          & 160ms         & 320ms         & 400ms         & 80ms          & 160ms         & 320ms         & 400ms         \\ \hline
Res. Sup.          & 35.4          & 57.3          & 76.3          & 87.7          & 34.5          & 63.4          & 124.6         & 142.5         & 38.0          & 69.3          & 115.0         & 126.7         & 36.1          & 69.1          & 130.5         & 157.1         \\
DMGNN              & 13.1          & 24.6          & 64.7          & 81.9          & 23.3          & 50.3          & 107.3         & 132.1         & 12.5          & 25.8          & 48.1          & 58.3          & 15.3          & 29.3          & 71.5          & 96.7          \\
LTD                & 9.0           & 19.9          & 43.4          & \underline{53.7}          & 18.7          & 38.7          & 77.7          & 93.4          & 10.2          & 21.0          & 42.5          & 52.3          & 13.7          & 29.9          & \underline{66.6}          & \underline{84.1}          \\
MSR                & \underline{8.6}           & \underline{19.7}          & \underline{43.3}          & 53.8          & \underline{16.5}          & \underline{37.0}          & \underline{77.3}          & \underline{93.4}          & \underline{10.1}          & \underline{20.7}          & \underline{41.5}          & \underline{51.3}          & \underline{12.8}          & \underline{29.4}          & 67.0          & 85.0          \\
Ours                & \textbf{7.2}  & \textbf{17.6} & \textbf{40.9} & \textbf{51.5} & \textbf{15.2} & \textbf{34.1} & \textbf{71.6} & \textbf{87.1} & \textbf{8.3}  & \textbf{18.3} & \textbf{38.7} & \textbf{48.4} & \textbf{10.7} & \textbf{25.7} & \textbf{60.0} & \textbf{76.6} \\ \hline
scenarios          & \multicolumn{4}{c|}{purchases}                                 & \multicolumn{4}{c|}{sitting}                                   & \multicolumn{4}{c|}{sittingdown}                               & \multicolumn{4}{c}{takingphoto}                               \\ \hline
millisecond        & 80ms          & 160ms         & 320ms         & 400ms         & 80ms          & 160ms         & 320ms         & 400ms         & 80ms          & 160ms         & 320ms         & 400ms         & 80ms          & 160ms         & 320ms         & 400ms         \\ \hline
Res. Sup.          & 36.3          & 60.3          & 86.5          & 95.9          & 42.6          & 81.4          & 134.7         & 151.8         & 47.3          & 86.0          & 145.8         & 168.9         & 26.1          & 47.6          & 81.4          & 94.7          \\
DMGNN              & 21.4          & 38.7          & 75.7          & 92.7          & 11.9          & 25.1          & 44.6          & \textbf{50.2}          & 15.0          & 32.9          & 77.1          & 93.0          & 13.6          & 29.0          & 46.0          & 58.8          \\
LTD                & 15.6          & 32.8          & 65.7          & \underline{79.3}          & 10.6          & \underline{21.9}          & 46.3          & 57.9          & 16.1          & \underline{31.1}          & \underline{61.5}          & \underline{75.5}          & 9.9           & \underline{20.9}          & 45.0          & 56.6          \\
MSR                & \underline{14.8}          & \underline{32.4}          & 66.1          & 79.6          & \underline{10.5}          & 22.0          & \underline{46.3}          & 57.8          & \underline{16.1}          & 31.6          & 62.5          & 76.8          & \underline{9.9}           & 21.0          & \underline{44.6}          & \underline{56.3}          \\
Ours                & \textbf{12.5} & \textbf{28.7} & \textbf{60.1} & \textbf{73.3} & \textbf{8.8}  & \textbf{19.2} & \textbf{42.4} & \underline{53.8} & \textbf{13.9} & \textbf{27.9} & \textbf{57.4} & \textbf{71.5} & \textbf{8.4}  & \textbf{18.9} & \textbf{42.0} & \textbf{53.3} \\ \hline
scenarios          & \multicolumn{4}{c|}{waiting}                                   & \multicolumn{4}{c|}{walkingdog}                                & \multicolumn{4}{c|}{walkingtogether}                           & \multicolumn{4}{c}{average}                                       \\ \hline
millisecond        & 80ms          & 160ms         & 320ms         & 400ms         & 80ms          & 160ms         & 320ms         & 400ms         & 80ms          & 160ms         & 320ms         & 400ms         & 80ms          & 160ms         & 320ms         & 400ms         \\ \hline
Res. Sup.          & 30.6          & 57.8          & 106.2         & 121.5         & 64.2          & 102.1         & 141.1         & 164.4         & 26.8          & 50.1          & 80.2          & 92.2          & 34.7          & 62.0          & 101.1         & 115.5         \\
DMGNN              & 12.2          & 24.2          & 59.6          & 77.5          & 47.1          & 93.3          & 160.1         & 171.2         & 14.3          & 26.7          & 50.1          & 63.2          & 17.0          & 33.6          & 65.9          & 79.7          \\
LTD                & 11.4          & 24.0          & 50.1          & 61.5          & 23.4          & 46.2          & 83.5          & 96.0          & \underline{10.5}          & 21.0          & 38.5          & 45.2          & 12.7          & 26.1          & 52.3          & 63.5          \\
MSR                & \underline{10.7}          & \underline{23.1}          & \underline{48.3}          & 59.2          & \underline{20.7}          & \underline{42.9}          & \underline{80.4}          & \underline{93.3}          & 10.6          & \underline{20.9}          & \underline{37.4}          & \underline{43.9}          & \underline{12.1}          & \underline{25.6}          & \underline{51.6}          & \underline{62.9}          \\

Ours                & \textbf{8.9}  & \textbf{20.1} & \textbf{43.6} & \textbf{54.3} & \textbf{18.8} & \textbf{39.3} & \textbf{73.7} & \textbf{86.4} & \textbf{8.7}  & \textbf{18.6} & \textbf{34.4} & \textbf{41.0} & \textbf{10.3}  & \textbf{22.7}  & \textbf{47.4}  & \textbf{58.5} \\ \hline
\end{tabular}
}
\label{human3.6_shortterm}
\end{table*}

\begin{table*}[ht]
\caption{Comparisons of long-term prediction on Human3.6M. Results at 560ms and 1000ms in the future are shown.}
\vspace{-0.3cm}
\renewcommand\arraystretch{0.9}
\resizebox{\textwidth}{!}{
\begin{tabular}{c|cc|cc|cc|cc|cc|cc|cc|cc} \hline
\textbf{scenarios} & \multicolumn{2}{c|}{walking}    & \multicolumn{2}{c|}{eating}     & \multicolumn{2}{c|}{smoking}     & \multicolumn{2}{c|}{discussion}  & \multicolumn{2}{c|}{directions} & \multicolumn{2}{c|}{greeting}    & \multicolumn{2}{c|}{phoning}         & \multicolumn{2}{c}{posing}      \\ \hline
millisecond        & 560ms         & 1000ms         & 560ms         & 1000ms         & 560ms          & 1000ms         & 560ms          & 1000ms         & 560ms         & 1000ms         & 560ms          & 1000ms         & 560ms            & 1000ms           & 560ms          & 1000ms         \\ \hline
Res. Sup.          & 81.7          & 100.7          & 79.9          & 100.2          & 94.8           & 137.4          & 121.3          & 161.7          & 110.1         & 152.5          & 156.1          & 166.5          & 141.2            & 131.5            & 194.7          & 240.2          \\
DMGNN              & 73.4          & 95.8           & 58.1          & 86.7           & 50.9           & 72.2           & 81.9           & 138.3          & 110.1         & 115.8          & 152.5          & 157.7          & 78.9             & \textbf{98.6}             & 163.9          & 310.1          \\
LTD                & 54.1          & 59.8           & 53.4          & 77.8           & 50.7           & 72.6           & 91.6           & 121.5          & \underline{71.0}          & 101.8          & \underline{115.4}          & 148.8          & 69.2             & 103.1            & \underline{114.5}          & \underline{173.0}          \\
MSR                & \underline{52.7}          & \underline{63.0}           & \underline{52.5}          & \underline{77.1}           & \underline{49.5}           & \underline{71.6}           & \underline{88.6}           & \textbf{117.6} & 71.2          & \underline{100.6}          & 116.3          & \underline{147.2}          & \underline{68.3}             & 104.4            & 116.3          & 174.3          \\
Ours                & \textbf{48.1} & \textbf{56.4}  & \textbf{51.1} & \textbf{76.0}  & \textbf{46.5}  & \textbf{69.5}  & \textbf{87.1}  & \underline{118.2}          & \textbf{69.3} & \textbf{100.4} & \textbf{110.2} & \textbf{143.5} & \textbf{65.9}    & \underline{102.7}   & \textbf{106.1} & \textbf{164.8} \\ \hline
scenarios          & \multicolumn{2}{c|}{purchases}  & \multicolumn{2}{c|}{sitting}    & \multicolumn{2}{c|}{sittingdown} & \multicolumn{2}{c|}{takingphoto} & \multicolumn{2}{c|}{waiting}    & \multicolumn{2}{c|}{walkingdog}  & \multicolumn{2}{c|}{walkingtogether} & \multicolumn{2}{c}{average}     \\ \hline
millisecond        & 560ms         & 1000ms         & 560ms         & 1000ms         & 560ms          & 1000ms         & 560ms          & 1000ms         & 560ms         & 1000ms         & 560ms          & 1000ms         & 560ms            & 1000ms           & 560ms          & 1000ms         \\ \hline
Res. Sup.          & 122.7         & 160.3          & 167.4         & 201.5          & 205.3          & 277.6          & 117.0          & 143.2          & 146.2         & 196.2          & 191.3          & 209.0          & 107.6            & 131.1            & 97.6           & 130.5          \\
DMGNN              & 118.6         & 153.8          & \textbf{60.1}          & \textbf{104.9}          & 122.1          & 168.8          & 91.6           & 120.7          & 106.0         & 136.7          & 194.0          & 182.3          & 83.4             & 115.9            & 103.0          & 137.2          \\
LTD                & 102.0         & 143.5          & 78.3          & 119.7          & \underline{100.0}          & \underline{150.2}          & \underline{77.4}           & \underline{119.8}          & 79.4          & 108.1          & 111.9          & 148.9          & 55.0             & \underline{65.6}             & 81.6           & 114.3          \\
MSR                & \underline{101.6}         & \underline{139.2}          & 78.2          & 120.0          & 102.8          & 155.5          & 77.9           & 121.9          & \underline{76.3}          & \underline{106.3}          & \underline{111.9}          & \underline{148.2}          & \underline{52.9}             & 65.9             & \underline{81.1}           & \underline{114.2}          \\

Ours                & \textbf{95.3} & \textbf{133.3} & \underline{74.4} & \underline{116.1} & \textbf{96.7}  & \textbf{147.8} & \textbf{74.3}  & \textbf{118.6} & \textbf{72.2} & \textbf{103.4} & \textbf{104.7} & \textbf{139.8} & \textbf{51.9}    & \textbf{64.3}    & \textbf{76.9}  & \textbf{110.3} \\ \hline

\end{tabular} 
}
\label{Human3.6long-term}
\end{table*}

\section{Experiments}

\subsection{Datasets}

\textbf{Human3.6M}\footnote{The authors Tiezheng Ma and Yongwei Nie signed the license and produced all the experimental results in this paper. Meta did not have access to the Human3.6M dataset.}~\cite{ionescu2013human3} has 15 types of actions performed by 7 actors (S1, S5, S6, S7, S8, S9, and S11). Each pose has 32 joints in the format of exponential map. We convert them to 3D coordinates and angle representations, and discard 10 redundant joints. The global rotations and translations of poses are excluded. The frame rate is downsampled from 50fps to 25fps. S5 and S11 are used for testing and validation respectively, while the remaining are used for training.

\textbf{CMU-MoCap} has 8 human action categories. Each pose contains 38 joints in the format of exponential map which are also converted to 3D coordinates and angle representations. The global rotations and translations of the poses are excluded too. Following \cite{mao2019learning,dang2021msr}, we keep 25 joints and discard the others. The division of training and testing datasets is also the same as \cite{mao2019learning, dang2021msr}.

\textbf{3DPW} \cite{von2018recovering} is a challenging dataset containing human motion captured from both indoor and outdoor scenes. The poses in this dataset are represented in the 3D space. Each pose contains 26 joints and 23 of them are used (the other 3 are redundant).

\begin{table}[!t]
\huge
\caption{CMU-MoCap: comparisons of average prediction errors.}
\vspace{-0.3cm}
\resizebox{1.0\columnwidth}{!}{
\small
\begin{tabular}{c|cccccc} \hline
millisecond  & 80ms               & 160ms             & 320ms             & 400ms             & 560ms               & 1000ms              \\ \hline
Res. Sup. & 24.0               & 43.0              & 74.5              & 87.2              & 105.5               & 136.3               \\
DMGNN        & 13.6               & 24.1              & 47.0              & 58.8              & 77.4                & 112.6               \\
LTD          & 9.3                & 17.1              & 33.0              & 40.9              & 55.8                & 86.2                \\
MSR          & \underline{8.1}                & \underline{15.2}              & \underline{30.6}              & \underline{38.6}              & \underline{53.7}                & \underline{83.0}                \\
Ours          & \textbf{7.6} & \textbf{14.3} & \textbf{29} & \textbf{36.6} & \textbf{50.9} & \textbf{80.1} \\ \hline
\end{tabular}
}
\label{table:cmu_mocap}
\end{table}

\subsection{Comparison Settings}

\begin{table}[!t]
\begin{center}
\caption{3DPW: comparisons of average prediction errors.}
\vspace{-0.3cm}
\label{table-3DPW}
\renewcommand\arraystretch{1.0}
\resizebox{1.0\columnwidth}{!}{
\small
\begin{tabular}{c|ccccc} \hline
millisecond      & 200ms         & 400ms         & 600ms         & 800ms         & 1000ms         \\ \hline
Res. Sup.    & 113.9         & 173.1         & 191.9         & 201.1         & 210.7          \\
DMGNN  &37.3  &67.8  &94.5  &109.7  &123.6 \\
LTD       & \underline{35.6}          & \underline{67.8}          & \underline{90.6}          & \underline{106.9}         & \underline{117.8}          \\
MSR        & {37.8}          & {71.3}          & {93.9}          & {110.8}          & {121.5}          \\
Ours        & \textbf{29.3}& \textbf{58.3} & \textbf{79.8} & \textbf{94.4}  & \textbf{104.1} \\ \hline
\end{tabular}
}
\end{center}
\end{table}

\begin{table}[!t]
\center
\caption{Time and mode size comparisons.}
\vspace{-0.3cm}
\huge
\resizebox{1.0\columnwidth}{!}{
\small
\begin{tabular}{c|c|c|c} \hline
Method      & Train(Per batch) & Test(Per batch) & Model Size \\ \hline
DMGNN \cite{li2020dynamic}      & 473ms            & 85ms            & 46.90M         \\ \hline
LTD \cite{mao2019learning}        & 114ms            & 30ms            & 2.55M          \\ \hline
MSR \cite{dang2021msr}        & 191ms            & 57ms            & 6.30M          \\ \hline
Our         & 145ms            & 43ms            & 1.74M          \\ \hline
\end{tabular}
}
\label{table:overhead}
\end{table}

\textbf{Evaluation Metrics.} We train and test on both coordinate and angle representations. Due to the space limit, we only show the results measured by 3D coordinates in this paper. The results on angle can be found in the supplemental material. We use the Mean Per Joint Position Error (MPJPE) as our evaluation metric for 3D errors, and use Mean Angle Error (MAE) for angle errors.

\textbf{Test Scope.} We note that the works of \cite{martinez2017human,li2020dynamic,mao2019learning} randomly take 8 samples per action for test, Mao \etal \cite{mao2020history} randomly take 256 samples per action, and Dang \etal \cite{dang2021msr} take all the samples for test. We follow Dang \etal \cite{dang2021msr} to test on the whole test dataset in this paper. The comparison results on the random 8 and 256 test sets are provided in the supplemental material.

\textbf{Lengths of Input and Output Sequences.} Following \cite{dang2021msr}, the input length is 10 and the output is 25 for Human3.6M and CMU-MoCap, respectively. Following \cite{mao2019learning}, the input are 10 poses and the output are 30 poses for 3DPW. 

\textbf{Implementation Details}
Our multi-stage framework contains $T=4$ stages. In each Encoder-Copy-Decoder prediction network, the encoder contains 1 GCB and the decoder contains 2 GCBs. The framework contains 12 GCBs in total. We employ Adam as the solver. The learning rate is initially 0.005 and multiplied by $0.96$ after each epoch. The model is trained for 50 epochs with batchsize of 16. The devices we used are an NVIDIA RTX 2060 GPU and an AMD Ryzen 5 3600 CPU. For more implementation details, please refer to the supplemental material.

\subsection{Comparisons with previous approaches}

We compare our method with Res. Sup. \cite{martinez2017human}, DMGNN \cite{li2020dynamic}, LTD \cite{mao2019learning}, and MSR \cite{dang2021msr} on these three datasets~\footnote{We strictly comply with the agreement of using all the datasets for non-commercial research purpose only.}. Res. Sup. is an early RNN based approach. DMGNN uses GCN to extract features and RNN for decoding. LTD relies on GCN totally and performs the prediction in the frequency domain. MSR is a recent method executing LTD in a multiscale fashion. All these methods are previous state-of-the-arts which release their code publicly. For fair comparison, we use their pre-trained models or re-train the models using their default hyper-parameters.

\begin{figure*}[!t]
    \centering
    \includegraphics[width=\textwidth]{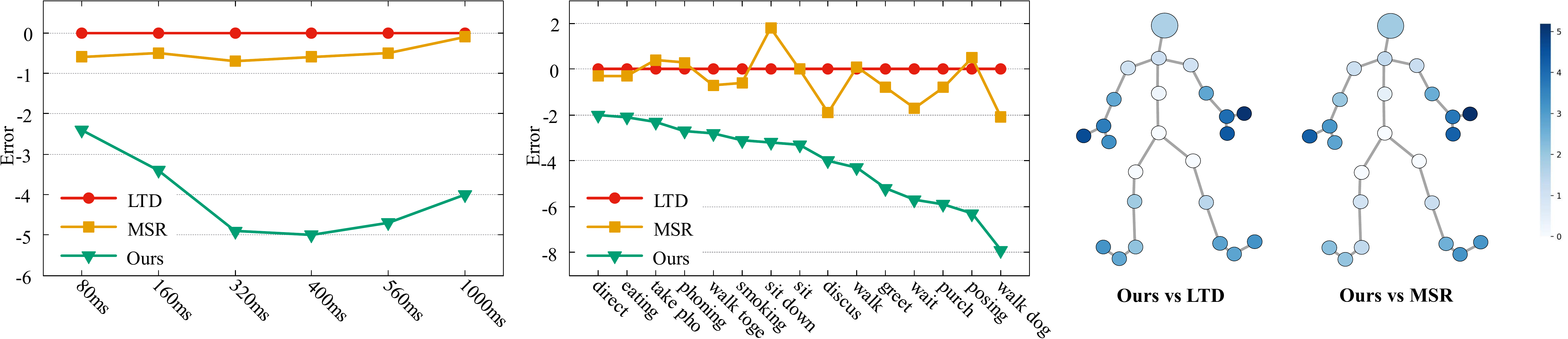} \\
    (a)\hspace{5.0cm}(b)\hspace{5.0cm}(c)\\
    \vspace{-0.3cm}
    \caption{Advantage analysis (Human3.6M). (a) The advantage of our method is most significant at 400ms. (b) The advantage of our method is most significant for the action of ``walking dog''. (c) The advantage per joint is illustrated. The darker the color, the greater the advantage of our method. }
    \label{fig:advantage-analysis}
    \vspace{-0.5cm}
\end{figure*}

\begin{figure} [!t]
    \centering
    \includegraphics[width=\columnwidth]{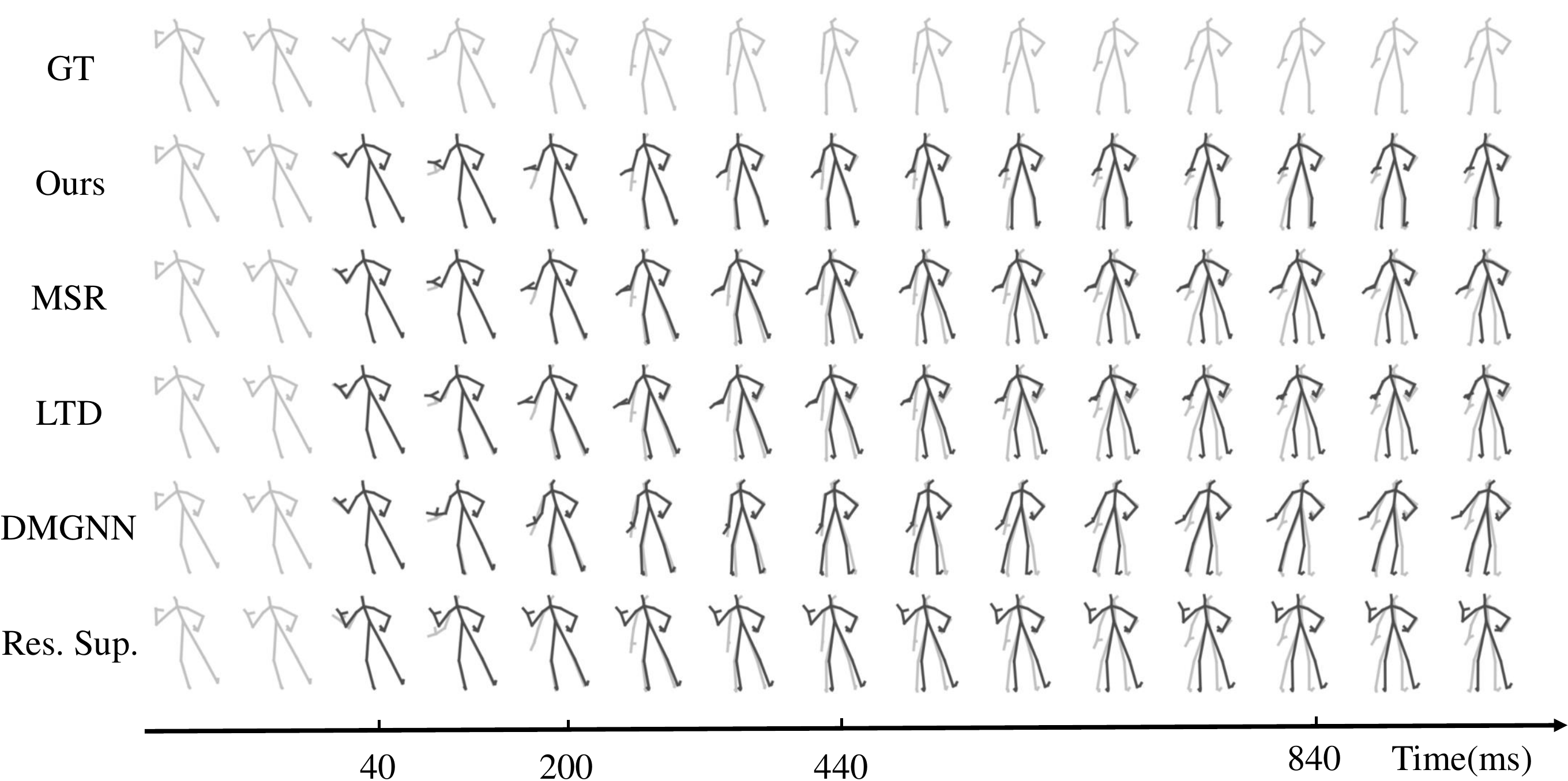} \\
    \vspace{-0.3cm}
    \caption{Visualization of predicted poses of different methods on a sample of Human3.6M.}
    \label{fig:qualitative}
\end{figure}

\textbf{Human3.6M.} Table \ref{human3.6_shortterm} shows the quantitative comparisons of short-term prediction (less than 400ms) on Human3.6M between our method and the above four approaches. Table \ref{Human3.6long-term} shows the comparisons of long-term prediction (more than 400ms but less than 1000ms) on Human3.6M. In most cases, our results are better than those of the compared methods. We show and compare the performance of different methods by statistics in Figure \ref{fig:advantage-analysis}. In Figure \ref{fig:advantage-analysis} (a) and (b), we treat LTD as the baseline, and subtract the prediction errors of MSR and our method from those of LTD. In (a), the relative average prediction errors with respect to LTD at every future timestamp are plotted. As can be seen, MSR is better than LTD, while our method is much better than MSR. Our advantage is the most significant at 400ms. In (b), the relative average prediction errors with respect to LTD for every action category are plotted. The advantage of our method compared with LTD and MSR is large, and for the action of ``walking dog'' the advantage is the most significant. In (c), we plot the advantage per joint of our method over LTD and MSR. The darker the color, the higher the advantage. As can be seen, our method achieves higher performance gain on limbs, especially on hands and feet. In Figure \ref{fig:qualitative}, we show an example of the predicted poses of different methods. With the increase of the forecast time, the result of our method becomes more and more better than those of the others.

\textbf{CMU-MoCap and 3DPW.} Table \ref{table:cmu_mocap} and Table \ref{table-3DPW} show the comparisons on CMU-MoCap and 3DPW respectively.  Due to space limit, we only show the average prediction errors at every timestamp. More detailed tables are provided in the supplementary material. On the two datasets, our method also outperforms the compared approaches. Especially, for the challenging dataset 3DPW, our advantage is very significant.

\textbf{Time and Model Size Comparisons.} As seen in Table \ref{table:overhead}, our model size is smaller than LTD (both models having 12 GCN blocks) as we use a smaller latent feature dimension than LTD (16 \vs 256). Our model is slightly slower than LTD due to the additional computations of intermediate losses and AAS, while faster than all the other methods.

\begin{table*}[!t]
\begin{center}
\caption{Ablations on architecture. Due to the space limit, please refer to the main text for the detailed descriptions of the experiments.}
\vspace{-0.3cm}
\label{table:ablation}
\resizebox{1.0\textwidth}{!}{
\small
\begin{tabular}{c|cccccccc|c} \hline
                &  80ms           & 160ms          & 320ms          & 400ms          & 560ms          & 720ms          & 880ms           & 1000ms          & average        \\ \hline
Single stage prediction & 11.95          & 24.47          & 49.69          & 60.94          & 79.56          & 93.93          & 105.86          & 113.41          & 67.48          \\

Without intermediate loss   & 11.42          & 24.02          & 49.73          & 60.94          & 79.49          & 93.45          & 105.12          & 112.42          & 67.07          \\
Supervised by GT at all stages &11.04 &23.49 &48.83 &59.89 &78.13 &92.20 &103.87 &111.46 &66.11 \\
Replacing Encoder-Copy-Decoder by LTD \cite{mao2019learning}      & 11.11          & 24.01          & 49.48          & 60.67          & 79.34          & 93.91          & 105.55          & 113.10          & 67.15          \\
Replacing S-DGCN, T-DGCN by ST-GCN \cite{yan2018spatial}       & 11.84          & 25.78          & 51.87          & 62.73          & 80.23          & 93.61          & 105.00          & 112.72          & 67.97          \\
Without ``Copy''  & 10.53          & 23.25          & 48.89          & 59.99          & 78.16          & 92.34          & 103.70          & 111.04          & 65.99          \\ \hline
Full model    & \textbf{10.33} & \textbf{22.74} & \textbf{47.45} & \textbf{58.46} & \textbf{76.91} & \textbf{91.20} & \textbf{102.77} & \textbf{110.31} & \textbf{65.02} \\ \hline
\end{tabular}
}
\end{center}
\vspace{-0.5cm}
\end{table*}

\begin{table}[!t]
\center
\caption{Ablations on ``Copy'' times and dimension.}
\vspace{-0.3cm}
\resizebox{\columnwidth}{!}{
\begin{tabular}{c|cc|c|cc} \hline
\textbf{Copy times} & \textbf{Error}               & \textbf{Model size} & \textbf{Copy dimension} & \textbf{Error} & \textbf{Model size} \\ \hline
No copy             & 65.99 & 1.06M               & Copy in channel         & 65.75          & 1.67M               \\
Copy once(Ours)     & \textbf{65.02} & 1.74M               & Copy in spatial         & 65.21          & 1.69M               \\
Copy three times    & 65.35                        & 3.28M               & Copy in temporal(Ours)  & \textbf{65.02} & 1.74M \\ \hline 
\end{tabular}
}
\label{table:ablation-copy}
\end{table}

\subsection{Ablation Analysis}
We conduct ablation studies to analyze our method in depth. All experimental results are obtained on Human3.6M.

\begin{table}[!t]
\caption{Comparisons between Gaussian filter and Accumulated Average Smoothing (AAS).}
\vspace{-0.3cm}
\huge
\resizebox{1.0\columnwidth}{!}{
\begin{tabular}{c|cccccc|c} \hline
                   & 80ms & 160ms & 320ms & 400ms & 560ms & 1000ms & average       \\ \hline
Gaussian-15 & 12.0 & 24.4  & 49.8  & 60.9  & 78.7  & 111.0  & 66.6          \\
Gaussian-21 & 11.5 & 23.7  & 48.8  & 60.0  & 78.5  & 112.2  & 66.5          \\ \hline
AAS & \textbf{10.3} & \textbf{22.7}  & \textbf{47.4}  & \textbf{58.5}  & \textbf{76.9}  & \textbf{110.3}  & \textbf{65.0} \\ \hline
\end{tabular}
}
\label{table:ablation-smmoothing-method}
\end{table}

\textbf{Architecture.} Several design choices contribute to the effectiveness of our method: (1) the multi-stage learning framework, (2) the intermediate supervisions, (2) the Encoder-Copy-Decoder prediction network, and (4) the ``Copy'' operator. Table \ref{table:ablation} shows the ablation experiments on different variants of the full model. The full model has 4 stages each containing 3 GCBs. There are 12 GCBs in total. The average prediction error is 65.02. (1) To show the effectiveness of ``multi-stage'', we test the case when $T=1$, \ie, there is only one Encoder-Copy-Decoder network which however has 12 GCBs with 6 GCBs in the encoder and 6 in the decoder. The prediction error becomes 67.48 which is a very large performance drop. (2) We use $T=4$ stages but remove the losses imposed on the intermediate outputs. The prediction error becomes 67.07, demonstrating the necessity of the intermediate supervisions. (3) In the third experiment, we use the ground truth (GT) to supervise all the intermediate outputs, which yields the prediction error of 66.11 on average. (4) We use LTD~\cite{mao2019learning} instead of the proposed Encoder-Copy-Decoder network to fulfill the task at each stage. The prediction error increases from 65.02 to 67.15. (5) We replace S-DGCN and T-DGCN by ST-GCN \cite{yan2018spatial}. The prediction error drastically increases from 65.02 to 67.97. (6) Finally, we remove the ``Copy'' operator in the middle of the Encoder-Copy-Decoder network, while yields a slightly increase of the prediction error from 65.02 to 65.99.

\begin{figure}[t]
    \centering
    \includegraphics[width=0.43\columnwidth]{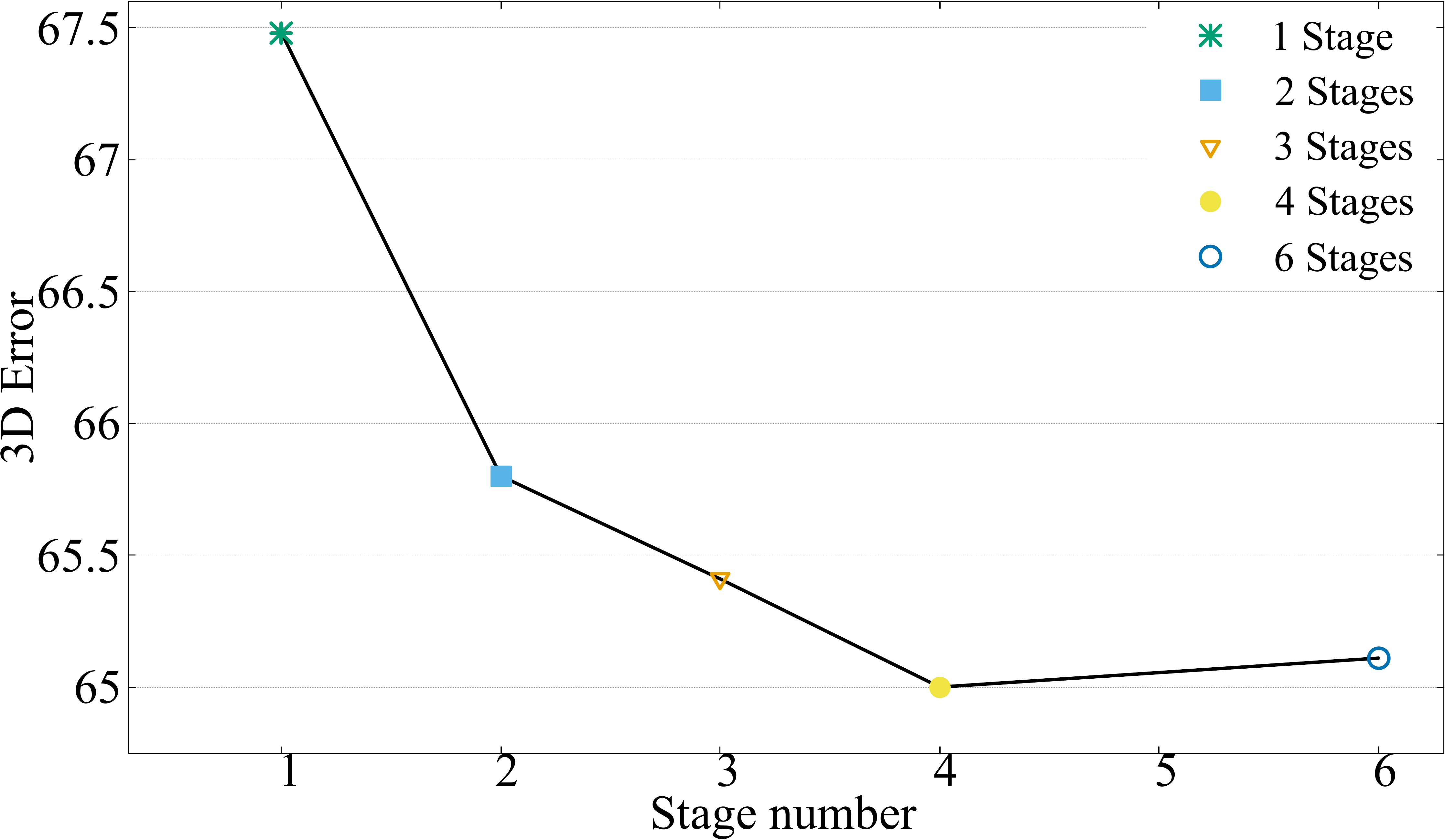}
    \includegraphics[width=0.52\columnwidth]{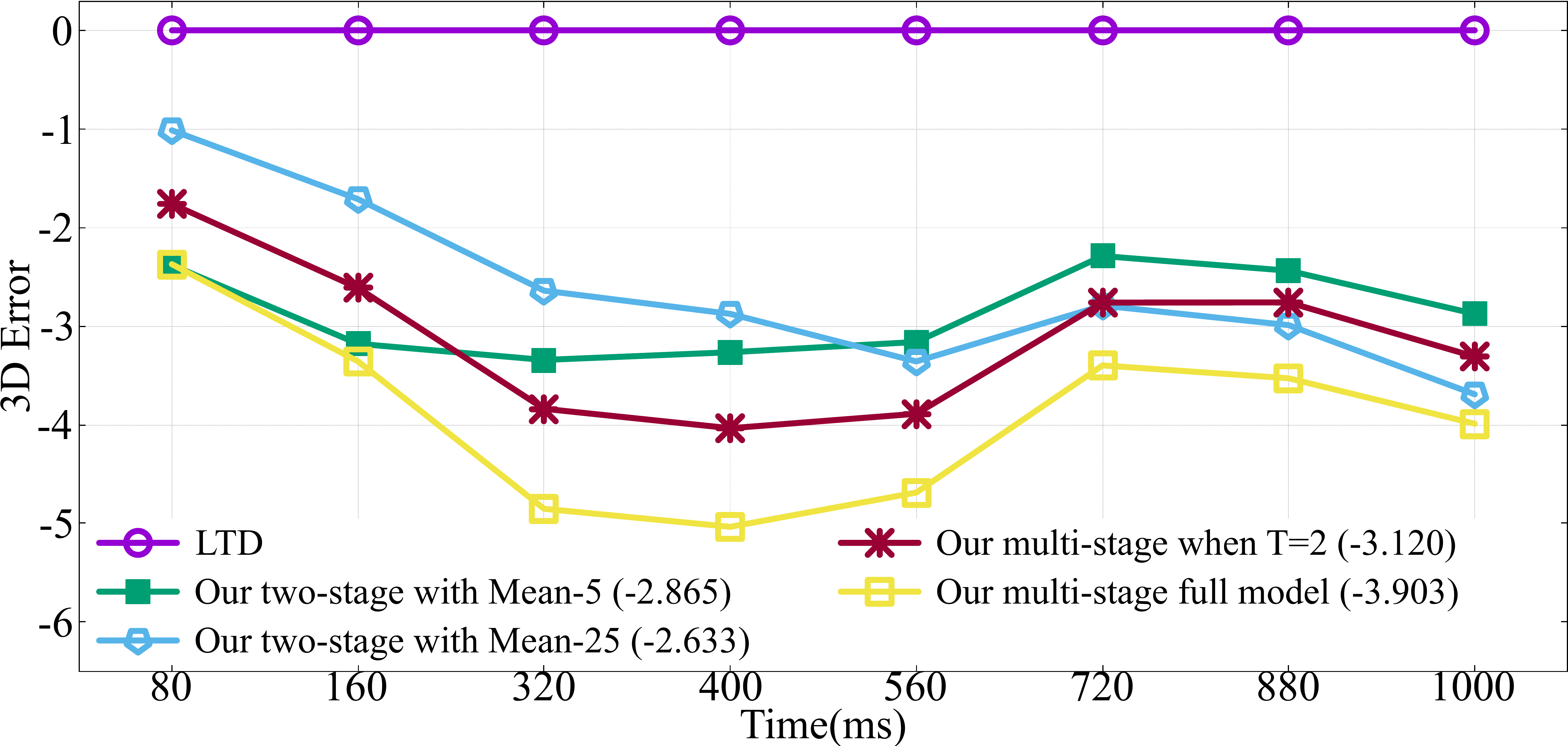}\\
    (a) \hspace{3.2cm} (b)
    \vspace{-0.3cm}
    \caption{(a) Ablation on the number of stages. The best result is achieved when $T=4$. (b) Comparisons between using Mean and AAS as intermediate target. LTD~\cite{mao2019learning} is the baseline. We subtract the prediction errors of LTD from those of the compared models. ``Our two-stage with Mean-$x$'' means using ``Mean-$x$'' as the intermediate target in the two-stage network. ``Our multi-stage'' means using the smoothed results by AAS as the intermediate targets.}
    \label{fig:ablation-stage-number}
\end{figure}

\textbf{Number of stages.} In Figure \ref{fig:ablation-stage-number} (a), we conduct ablations about $T$ from 1 to 6. For different $T$, the corresponding frameworks all contain 12 GCBs distributed in each stage network evenly. For example, if $T=3$, there will be 4 GCBs in each stage network. The experiments tell that the best performance is obtained when $T=4$.

\textbf{Direction and number of ``Copy''.} In the default setting of the Encoder-Copy-Decoder network, we copy the output of the encoder just one time and paste it along the temporal direction. In Table \ref{table:ablation-copy}, we conduct ablation studies on the number of copying and the direction of pasting. As can be seen, copying once or three times is better than not copying. But copying three times does not bring more performance gain than copying once. Copying once along the spatial dimension, the channel dimension and the temporal dimension are all better than not copying, while copying along the temporal dimension yields the best result.

\textbf{AAS \vs Gaussian filter.} In Table \ref{table:ablation-smmoothing-method}, we compare between Accumulated Average Smoothing (AAS) and Gaussian filter. ``Gaussian-$x$'' means the filtering window size is $x$. It can be seen that AAS performs better than the two Gaussian filters.

\textbf{AAS \vs Mean.} Recall that for our two-stage framework, \ie, the one shown in Figure \ref{fig:teaser} (b), we can use Mean-$x$ as the intermediate target. For the same framework, we can also use $S^{L-1}_{1:L}$ as the intermediate target. We call the two schemes ``Our two-stage with Mean-$x$'' and ``Our multi-stage when $T=2$'', respectively, and compare between them in Figure~\ref{fig:ablation-stage-number} (b). As can be seen, ``Our multi-stage when $T=2$'' is better than both ``Our two-stage with Mean-$5$'' and ``Our two-stage with Mean-$25$'', which demonstrate that the smoothed result by AAS is better than the global mean of the future poses when used as the intermediate target. ``Our multi-stage full model'' when $T=4$ achieves even better results.

\subsection{Limitations and Future Works}
Our method has two limitations: (1) The average prediction of LTD~\cite{mao2019learning} is 68.08. Ours is 65.02. In contrast, ``LTD with Mean-25'' in Figure~\ref{fig:teaser} (a) is 29.76. We still have much room to reduce the absolute prediction error. In the future, one  can investigate more effective intermediate targets. (2) Our method requires a set of poses as input, while in real applications the poses may be occluded. How to deal with incomplete observations is worthy of further study.

\section{Conclusion}
We have presented a multi-stage human motion prediction framework. The key to the effectiveness of the framework is that we decompose the originally difficult prediction task into many subtasks, and ensure each subtask is simple enough. We achieve this by taking the recursively smoothed versions of the target pose sequence as the prediction targets of the subtasks. The adopted Accumulated Average Smoothing strategy guarantees that the smoothest intermediate target approaches to the last observed data, and the intermediate target of the current stage is a good guess of the next stage. Besides that, we have proposed the novel Encoder-Copy-Decoder prediction network, the S-DGCN and T-DGCN of which can extract spatiotemporal features effectively while the ``Copy'' operator further enhances the capability of the decoder. We have conducted extensive experiments and analysis demonstrating the effectiveness and advantages of our method.

\section*{Acknowledgement}
This research is sponsored by Prof. Yongwei Nie's and Prof. Guiqing Li's National Natural Science Foundation of China (62072191, 61972160), and their Natural Science Foundation of Guangdong Province (2019A1515010860, 2021A1515012301).

{\small
\bibliographystyle{ieee_fullname}
\bibliography{PaperForReview}
}

\end{document}